\documentclass[10pt,twocolumn,letterpaper]{article}

\usepackage{wacv}

\usepackage{graphicx}
\usepackage{booktabs}
\usepackage{multirow}
\usepackage{amsmath}
\usepackage{amssymb}
\usepackage{array}
\usepackage{makecell}
\usepackage{placeins}
\usepackage{indentfirst}
\usepackage{tikz}
\usetikzlibrary{arrows.meta,positioning}

\newcommand{\method}{DiTGaze\xspace}
\newcommand{\degrees}{^{\circ}}

\definecolor{wacvblue}{rgb}{0.21,0.49,0.74}
\usepackage[pagebackref,breaklinks,colorlinks,allcolors=wacvblue]{hyperref}

\def\wacvPaperID{3142}
\def\confName{WACV}
\def\confYear{2027}

\title{DiTGaze: Geometry-Conditioned Diffusion Augmentation for Few-Shot Gaze Estimation}

\author{Abiram Panchalingam \qquad Indu P. Bodala \qquad Stuart E. Middleton\\
School of Electronics and Computer Science, University of Southampton, UK}

\begin{document}
\maketitle

\begin{abstract}
Accurate person-specific gaze estimation benefits from labeled user data, but collecting images that densely cover gaze directions is burdensome. We present \textbf{DiTGaze}, a calibration-efficient augmentation system that follows a two-stream 3D Gaussian Splatting backbone and reformulates its convolutional rendering stage as a geometry-conditioned latent Diffusion Transformer (DiT). A learnable projection and frozen VAE bridge rasterized Gaussian features to latent tokens, while timestep and identity/expression conditioning are integrated through Adaptive Layer Normalization. Intermediate yaw--pitch sampling densifies gaze supervision, and an orthogonality constraint reduces interference between gaze, pose, and expression. On ETH-XGaze, DiTGaze reduces generator-level gaze error by 4.1\% relative to GazeGaussian. In our runs of the expanded downstream protocol, it ranks first at every calibration budget from one to nine real images; the repeated primary comparison at $N=9$ reduces subject-level estimator error by $0.73\degrees$ and improves 14 of 15 identities ($p<0.001$). These results connect the proposed generator changes to statistically supported few-shot personalization gains.
\end{abstract}

\section{Introduction}
\label{sec:intro}

Gaze estimation supports attentive interfaces, driver monitoring, assistive systems, and learning technologies. In engagement-aware tutoring and e-learning, small gaze differences can help distinguish attention to relevant content from distraction, making low angular error important rather than merely cosmetic~\cite{Sharma2020EyeTrackingLearning}. Yet gaze estimators are sensitive to identity, illumination, camera, and head-pose shifts. Person-specific estimator fine-tuning can reduce this gap, but acquiring dense gaze coverage from every user is intrusive and time-consuming. An attractive alternative is to redirect gaze in a few observed images, producing controlled synthetic views that expand coverage while preserving the person's identity and non-gaze attributes.

This is a demanding data-generation problem. Small errors in iris position alter the target label, while changes to identity, expression, or head pose introduce unwanted supervision. Image-space warping and 2D generative approaches~\cite{Ganin2016DeepWarp,He2019PhotoRealistic,Zheng2020STED} do not explicitly represent eye geometry. GazeNeRF~\cite{Ruzzi2022GazeNeRF} introduced 3D-aware two-stream rendering, but volumetric sampling is costly. GazeGaussian~\cite{Wei2025GazeGaussian} made this pipeline more efficient through separate face and eye Gaussian streams, an Eye Rotation Field for explicit gaze control, and an expression-guided convolutional renderer. We follow its two-stream 3DGS backbone and redesign how the controlled representation is supervised, factorized, and rendered.

We address the application question: \emph{when can geometry-conditioned redirection turn a small calibration set into useful supervision for person-specific gaze estimation?} We introduce \method, an offline augmentation system with three coordinated enhancements to the 3DGS pipeline. First, an Intermediate Gaze Sampler supplies yaw--pitch targets beyond the discrete training pairs, directly expanding the directions learned by the Eye Rotation Field. Second, an orthogonality loss regularizes the deformation embeddings before Gaussian transformation, reducing expression-related interference with gaze and pose. Third, we replace the expression-guided convolutional renderer with a latent DiT~\cite{Peebles2022DiT}. This is not a drop-in transformer substitution: a learnable projection adapts the rasterizer's 32-channel feature map to a frozen VAE, the resulting four-channel latent is patchified, and timestep plus identity/expression information conditions denoising through AdaLN. The Eye Rotation Field determines the requested geometry before the DiT reconstructs coherent identity, texture, and illumination.

Using redirected images for estimator fine-tuning is established~\cite{Ruzzi2022GazeNeRF,Wei2025GazeGaussian}; synthetic augmentation alone is therefore not our novelty claim. Our systems contribution is an integrated, calibration-efficient formulation and a controlled assessment that connects target-label fidelity, identity preservation, angular coverage, non-gaze preservation, and offline cost to task-level utility. A visually convincing redirector can still introduce label or identity errors that hurt personalization, and a single calibration selection cannot establish that a small difference between generators is robust.

Our evaluation separates generator quality from application utility. Experiment~1 measures redirection accuracy, non-gaze preservation, identity, and perceptual fidelity. Experiment~2 reports our runs of the enhanced one-to-nine-shot downstream protocol, in which each redirector expands the same person-specific real set to 200 fine-tuning samples. \method ranks first at every calibration budget. A separate, repeated $N=9$ comparison aggregates five calibration draws and three estimator seeds, reducing mean subject-level error by $0.73\degrees$ relative to GazeGaussian. Downstream component, real-shot, synthetic-count, and domain tests further examine when these gains persist or saturate.

To make the scope of novelty explicit, we separate changes to the generator from contributions in its application and evaluation.

\noindent\textbf{Architectural and model-design contributions.}
\begin{itemize}[leftmargin=*,topsep=2pt,itemsep=1pt,parsep=0pt]
    \item A \textbf{geometry-conditioned latent DiT renderer} that enhances the 3DGS pipeline established by Gaussian Head Avatar and adapted for gaze control by GazeGaussian. A learned feature-map projection, frozen VAE, latent patch interface, and AdaLN conditioning integrate the diffusion timestep and identity/expression guidance while preserving explicit 3D gaze control.
    \item \textbf{Continuous gaze-space supervision and deformation regularization}: intermediate yaw--pitch sampling expands angular coverage beyond discrete training pairs, while orthogonally constrained deformation embeddings reduce expression-related interference with gaze and pose.
\end{itemize}

\noindent\textbf{Application and testing contributions.}
\begin{itemize}[leftmargin=*,topsep=2pt,itemsep=1pt,parsep=0pt]
    \item A complete \textbf{calibration-efficient gaze augmentation system} that converts sparse person-specific observations into controlled synthetic supervision and improves a downstream estimator under a fixed real-plus-synthetic data protocol.
    \item A \textbf{requirements-driven comparative assessment} spanning paired subject-level statistics, calibration and synthetic-data budgets, downstream component ablations, in- and cross-domain use, redirection quality, compute, and failure modes. In the repeated $N=9$ comparison, \method improves mean subject-level error by $0.73\degrees$ over GazeGaussian.
\end{itemize}

\section{Related work}
\label{sec:related}

\paragraph{Gaze estimation and synthetic augmentation.}
Cross-person and cross-domain variation remains a central obstacle in appearance-based gaze estimation. Large-scale datasets such as ETH-XGaze~\cite{xu2020ethxgaze}, MPIIFaceGaze~\cite{zhang2015mpiifacegaze,zhang2016fullfacegaze}, GazeCapture~\cite{krafka2016eyetracking}, and ColumbiaGaze~\cite{smith2013gazelocking} capture complementary viewing conditions, but collecting dense labels for a new user is still expensive. Recent work also studies difficult redirection pairs and fine angular changes~\cite{Park2023HardnessGaze}. Redirection provides targeted augmentation by changing the label-relevant gaze factor while retaining other properties of a real calibration image. Prior 3D redirectors already use generated samples in person-specific gaze-estimator fine-tuning~\cite{Ruzzi2022GazeNeRF,Wei2025GazeGaussian}; we extend this downstream task to compare redirectors under matched data and optimization controls.

Synthetic gaze supervision predates neural redirection: SynthesEyes and UnityEyes render labeled eye images across gaze, pose, and illumination~\cite{Wood2015SynthesEyes,Wood2016UnityEyes}. Yu \etal~\cite{Yu2019FewShotRedirection} use gaze-redirection synthesis for few-shot user adaptation; FAZE~\cite{Park2019FAZE} is a synthesis-free few-shot gaze estimator using rotation-aware representations and meta-learning; and Render-to-Adapt~\cite{Ge2026RenderToAdapt} uses a gaze-conditioned renderer for unsupervised adaptation from unlabeled user images. Our application contribution is the integrated generator and controlled requirements-driven assessment, rather than synthetic augmentation alone.

\paragraph{Gaze redirection.}
Early methods manipulate pixels through learned warping~\cite{Ganin2016DeepWarp,Shu2017EyeOpener} or gaze-specific adversarial generation~\cite{He2019PhotoRealistic}, while StarGAN~\cite{Choi2017StarGAN} translates facial attributes and expressions, not gaze. STED~\cite{Zheng2020STED} and ReDirTrans~\cite{Jin2023ReDirTrans} rotate learned latent factors, improving controllability but retaining an implicit 2D representation. Fine gaze-redirection learning interpolates feature representations according to pair hardness~\cite{Park2023HardnessGaze}; our sampler instead introduces targets in the explicit yaw--pitch space used by the 3D transformation field. GazeNeRF~\cite{Ruzzi2022GazeNeRF} separates face and eye neural fields and rotates the eye representation in 3D before volumetric compositing, building on neural radiance fields~\cite{Mildenhall2022NeRF}. GazeGaussian~\cite{Wei2025GazeGaussian} uses two-stream 3DGS and an Eye Rotation Field for explicit control and rasterization-based rendering. \method follows this backbone while modifying its target supervision, deformation regularization, and rendering interface.

\paragraph{Controllable head synthesis and diffusion.}
Parametric models such as FLAME~\cite{Li2017FLAME}, NeRF-based heads such as HeadNeRF~\cite{Hong2021HeadNeRF}, point-based avatars~\cite{Zheng2022PointAvatar}, and rigged Gaussian avatars~\cite{Qian2023GaussianAvatars,xu2024gaussian} provide increasingly controllable head geometry. Gaussian Head Avatar~\cite{xu2024gaussian} supplies the neutral-mesh Gaussian deformation framework later adapted for gaze-specific two-stream control, but general head-avatar methods are not designed to preserve precise gaze labels.

Diffusion models~\cite{Ho2020DDPM} traditionally use convolutional U-Nets~\cite{Ronneberger2015UNet}; latent diffusion~\cite{Rombach2021LDM} reduces cost by denoising a compressed representation. DiT~\cite{Peebles2022DiT} replaces the U-Net with self-attention over latent patches, and transformer denoisers have also been investigated for 3D shape~\cite{Mo2023DiT3D} and medical-image synthesis~\cite{Pan2023MedDiffusion}. Rather than applying diffusion as unconstrained face editing, we condition it on rasterized, explicitly transformed 3D Gaussians. Our regularizer follows the general goal of factorized representation learning~\cite{higgins2017betavae,kim2018factorvae}, but acts directly on the factor-specific deformation embeddings that drive the 3D representation.

\section{DiTGaze augmentation system}
\label{sec:method}

\subsection{Application workflow}
\label{sec:workflow}

We reserve \emph{training} for one-time population-level learning on the ETH-XGaze training identities. At test time, the \emph{calibration budget} $N$ is the number of labeled real images provided by an unseen subject, while the \emph{calibration set} contains those specific $N$ images and their gaze labels. These images are used in two test-time stages. First, \emph{generator identity adaptation} performs a 30 s subject-specific update of the identity-specific Gaussian attributes, deformation parameters, and identity code while keeping the trained global renderer fixed. The adapted generator then redirects each calibration image toward sampled target gaze directions. Second, \emph{estimator fine-tuning} optimizes the ResNet-50 gaze estimator using the calibration images alone or a fixed-size combination of real and generated images. We use \emph{personalization} to refer collectively to generator identity adaptation and estimator fine-tuning. Image generation is performed offline, prioritizing synthesis quality and gaze-label fidelity over live frame rate.

\begin{figure*}[t]
\centering
\begin{tikzpicture}[
    box/.style={draw=black!55, rounded corners=2pt, align=center, minimum height=9mm, text width=21.5mm, fill=green!5, font=\small},
    arr/.style={-{Latex[length=2mm]}, thick, draw=black!65},
    node distance=4mm
]
\node[box] (real) {$N=9$ labeled real\\calibration images};
\node[box, right=of real] (adapt) {30 s generator\\identity adaptation};
\node[box, right=of adapt] (targets) {controlled target\\gaze directions};
\node[box, right=of targets] (generate) {DiTGaze offline\\generation};
\node[box, right=of generate] (set) {9 real + 191 synthetic\\estimator fine-tuning set};
\node[box, right=of set] (tune) {fine-tune\\ResNet-50};
\draw[arr] (real) -- (adapt);
\draw[arr] (adapt) -- (targets);
\draw[arr] (targets) -- (generate);
\draw[arr] (generate) -- (set);
\draw[arr] (set) -- (tune);
\end{tikzpicture}
\caption{\textbf{Primary $N=9$ application protocol.} The nine labeled calibration images both adapt the generator and remain in each augmented fine-tuning set (9 real + 191 synthetic); real-only uses the nine real images. Evaluation uses 185 held-out images from the same unseen subject.}
\label{fig:application_workflow}
\end{figure*}

\begin{figure*}[t]
    \centering
    \includegraphics[width=\linewidth,trim=20 0 20 0,clip]{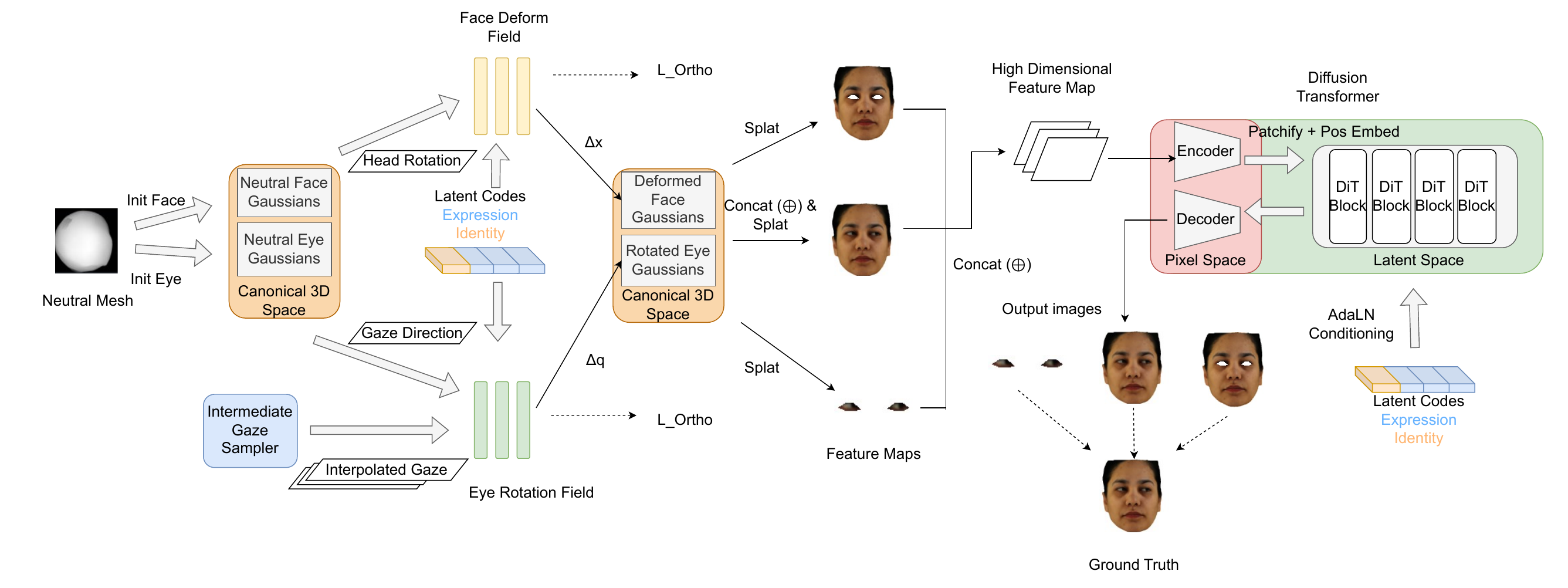}
    \caption{\textbf{DiTGaze generator.} Face and eye Gaussians form a two-stream 3DGS backbone. Pose and expression drive the Face Deform Field to produce offsets $\Delta x$, while sampled target gaze drives the Eye Rotation Field to produce rotations $\Delta q$. Orthogonality regularizes the deformation embeddings, and the transformed Gaussians are rasterized into feature maps for the latent DiT renderer, which uses AdaLN conditioning to synthesize the gaze-redirected image.}
    \label{fig:pipeline}
\end{figure*}

\Cref{fig:pipeline} summarizes our contributions to the 3DGS pipeline. We follow the two-stream 3DGS backbone for separate face deformation and explicit eye rotation, then modify three complementary stages: sampled yaw--pitch targets expand supervision, orthogonality regularizes the deformation embeddings, and a geometry-conditioned latent DiT replaces the expression-guided convolutional renderer. The first two changes strengthen the controlled Gaussian state before rasterization; the DiT then reconstructs identity, texture, and illumination without selecting the gaze direction.

\paragraph{Fixed-budget application formulation.}
For an unseen identity $u$, let $\mathcal{C}_u=\{(I_i,g_i)\}_{i=1}^{N}$ be the real calibration set, where $g_i$ is the corresponding 3D gaze label and $N=9$ in our primary paired comparison. The adapted generator $S_{\theta,u}$ maps a selected calibration view and requested direction $g'$ to a redirected image. With $s(j)$ selecting a real source index, we form
\begin{equation}
\begin{aligned}
\mathcal{A}_u={}&\mathcal{C}_u\cup
\{(S_{\theta,u}(I_{s(j)},g'_j),g'_j)\}_{j=1}^{B-N},\\[-1mm]
&|\mathcal{A}_u|=B,\qquad B=200,
\end{aligned}
\label{eq:application_set}
\end{equation}
and fine-tune the same pre-trained gaze estimator on $\mathcal{A}_u$ for every generator. A useful system must therefore satisfy more than visual realism: $g'_j$ must remain a reliable label, identity and non-gaze factors should be retained from $I_{s(j)}$, requested targets should add angular coverage, and end-to-end cost must fit offline personalization. \Cref{sec:experiments} measures generator-level requirements, and \Cref{sec:downstream} tests their combined effect through held-out estimator error.

\subsection{Two-stream 3D Gaussian backbone}

We follow the two-stream 3DGS backbone~\cite{Wei2025GazeGaussian}, whose neutral-mesh initialization and deformation machinery build on Gaussian Head Avatar~\cite{xu2024gaussian}. A neutral mesh is extracted from a signed-distance representation using DMTet~\cite{Shen2021DMTet,park2019deepsdf,gropp2020igr} and partitioned with facial landmarks. Face and eye masks are obtained with the same face-parsing pipeline as GazeGaussian~\cite{zllfaceparsing}. The mesh initializes canonical face and eye Gaussian sets $\mathcal{G}^{0}_{f}$ and $\mathcal{G}^{0}_{e}$.

In 3D Gaussian Splatting (3DGS), the head is represented by explicit, differentiable primitives rather than a continuously sampled volume. Using compact notation, each stream $r\!\in\!\{f,e\}$ contains
\[
\mathcal{G}^{0}_{r}=\{(\mathbf{p}^{r}_{k},\mathbf{f}^{r}_{k},
\mathbf{a}^{r}_{k})\}_{k=1}^{K_r},
\]
where $\mathbf{p}$ is the 3D center, $\mathbf{f}$ is an appearance feature, and $\mathbf{a}$ groups rotation, scale, and opacity. For a camera view, each Gaussian is projected to an elliptical image-space footprint; depth-ordered alpha compositing then accumulates the overlapping footprints into a rasterized feature map. This explicit splatting avoids the dense point sampling and ray integration used by NeRF. Face Gaussians permit anisotropic scale for non-rigid deformation, whereas the eye stream uses isotropic scale to better approximate rigid eyeball rotation.

The Face Deform Field predicts pose- and expression-conditioned updates,
\begin{equation}
\begin{aligned}
\Delta x ={}& \lambda_{\mathrm{expr}}
\Delta_{\mathrm{expr}}(\mathcal{G}^{0}_{f},\tau) \\
&+\lambda_{\mathrm{pose}}
\Delta_{\mathrm{pose}}(\mathcal{G}^{0}_{f},\gamma),\\[-1mm]
\mathcal{G}_{f} ={}& \mathcal{G}^{0}_{f}+\Delta x .
\end{aligned}
\label{eq:face}
\end{equation}
where $\tau$ and $\gamma$ are expression and head-pose codes. The landmark-aware weights $\lambda_{\mathrm{expr}}$ and $\lambda_{\mathrm{pose}}$ control their relative influence across the face. In this compact representation, $\Delta x$ denotes the predicted updates to the affected Gaussian state, including position and attribute changes; it is not a 2D image-space warp. The Eye Rotation Field independently predicts
\begin{equation}
\begin{aligned}
\Delta q ={}& \Delta_{\mathrm{expr}}(\mathcal{G}^{0}_{e},\tau)
+\Delta_{\mathrm{gaze}}(\mathcal{G}^{0}_{e},\phi),\\[-1mm]
\mathcal{G}_{e} ={}& \mathcal{G}^{0}_{e}+\Delta q .
\end{aligned}
\label{eq:eye}
\end{equation}
for target gaze $\phi$. Here $\Delta q$ compactly groups the expression-conditioned eye deformation and the gaze-conditioned positional, rotational, and appearance updates. The gaze branch approximates rigid eyeball motion, while the expression branch accounts for surrounding eye appearance such as eyelid deformation. The transformed sets are concatenated and splatted to a structured feature map $\mathcal{M}=\mathcal{R}(\mathcal{G}_{f}\cup\mathcal{G}_{e})$. This explicit path ties the intended gaze label to a 3D eye transformation instead of relying on an unconstrained image edit; the subsequent DiT refines the rasterized appearance but does not choose the gaze direction.

\subsection{Geometry-conditioned latent DiT}

Gaze redirection is local in the eyes but the rendered result must remain coherent with full-face identity, expression, illumination, and head geometry. We therefore replace GazeGaussian's U-Net-like Expression-Guided Neural Renderer with a globally attentive latent DiT~\cite{Peebles2022DiT}. This is not a drop-in decoder replacement: the Gaussian rasterizer produces a 32-channel feature map, whereas the frozen latent-diffusion VAE~\cite{Rombach2021LDM} expects an RGB-like input and the DiT operates on latent tokens. A learnable $1\!\times\!1$ projection, trained end-to-end with the renderer, first maps $\mathcal{M}$ to a 3-channel VAE-compatible representation. The VAE compresses it to four latent channels, after which patch embedding forms tokens and two-dimensional sinusoidal positions preserve spatial structure.

Six DiT blocks, each with eight attention heads, iteratively denoise the latent tokens. To retain the subject-specific guidance provided by GazeGaussian's expression-guided renderer while incorporating the diffusion state, we combine the timestep embedding with identity/expression codes in conditioning vector $c$. Adaptive Layer Normalization injects learned scale and shift into every block:
\begin{equation}
\operatorname{adaLN}(h,c)=\bigl(1+s(c)\bigr)\odot\operatorname{LN}(h)+b(c),
\label{eq:adaln}
\end{equation}
where $s(\cdot)$ and $b(\cdot)$ predict channel-wise scale and shift. Thus, adopting AdaLN required a new conditioning path matched to the Gaussian identity/expression codes and diffusion timestep, in addition to the feature-map-to-latent interface. A final projection and unpatchification reconstruct the latent map, which the frozen VAE decoder converts to the redirected image. The trainable projection, DiT, Gaussian attributes, and deformation fields are optimized jointly; the VAE weights remain fixed.

\subsection{Continuous gaze-space supervision}
\label{sec:sampler}

Training pairs cover a discrete subset of the continuous gaze space. Interpolating only network features need not respect angular geometry. Instead, our Intermediate Gaze Sampler creates yaw--pitch targets directly in the same 3D gaze coordinate system used by the Eye Rotation Field. For an $N_{\mathrm{grid}}\!\times\!N_{\mathrm{grid}}$ grid with angular range $R_{\mathrm{gaze}}$,
\begin{equation}
\begin{aligned}
g^{\mathrm{yaw}}_{i} &=-R_{\mathrm{gaze}}
 +\frac{2R_{\mathrm{gaze}}i}{N_{\mathrm{grid}}-1},\\[-1mm]
g^{\mathrm{pitch}}_{j} &=-R_{\mathrm{gaze}}
 +\frac{2R_{\mathrm{gaze}}j}{N_{\mathrm{grid}}-1}.
\end{aligned}
\label{eq:sampling}
\end{equation}
A mixed schedule combines these deterministic grid points with random samples. The grid guarantees repeatable angular coverage, while random targets prevent the controller from overfitting a fixed lattice and expose directions between grid cells. Early epochs emphasize center-biased directions for stable optimization; later epochs expand toward uniform coverage. Unlike feature-space interpolation~\cite{Park2023HardnessGaze}, the sampled vector directly conditions the Eye Rotation Field. It replaces the target-image gaze in the gaze-redirection loss, so the model receives functional angular supervision even when no pixel-aligned image exists for that exact intermediate direction. The image reconstruction terms are retained only where an observed target exists.

\subsection{Orthogonally constrained deformation}

Separate face and eye streams do not prevent their internal features from becoming correlated. We collect the first-layer embeddings of the gaze-, pose-, and expression-conditioned deformation MLPs, denoted $v_{\mathrm{gaze}}$, $v_{\mathrm{pose}}$, and $v_{\mathrm{expr}}$. Expression changes are the dominant appearance deformation, so we regularize its correlation with gaze and pose:
\begin{equation}
\mathcal{L}_{\mathrm{ortho}}=
w_1\left|\cos(v_{\mathrm{gaze}},v_{\mathrm{expr}})\right|
+w_2\left|\cos(v_{\mathrm{pose}},v_{\mathrm{expr}})\right|.
\label{eq:ortho}
\end{equation}
This acts before Gaussian transformation and rasterization. It is deliberately not an all-pairs disentanglement claim: gaze and head pose already use structurally separate transformation paths, whereas expression affects both face and eye appearance. The regularizer therefore targets expression-related interference with gaze and pose, without adding an auxiliary encoder.

The full objective combines region-aware image synthesis, functional gaze consistency, and deformation regularization:
\begin{equation}
\mathcal{L}=\lambda_I\mathcal{L}_I+\lambda_G\mathcal{L}_G
+\lambda_{\mathrm{ortho}}\mathcal{L}_{\mathrm{ortho}}.
\label{eq:total}
\end{equation}
For each supervised region $r\in\{\text{eye},\text{face},\text{head}\}$, the image term is
\begin{equation}
\begin{aligned}
\mathcal{L}^{(r)}_I={}&\|I^{(r)}_{\mathrm{gt}}-I^{(r)}_{\mathrm{out}}\|_1
+\lambda_{\mathrm{SSIM}}(1-\mathrm{SSIM})\\
&+\lambda_{\mathrm{perc}}\,
\mathcal{D}_{\mathrm{VGG}}(I^{(r)}_{\mathrm{gt}},I^{(r)}_{\mathrm{out}}),
\end{aligned}
\label{eq:image_loss}
\end{equation}
with $\mathcal{L}_I$ the weighted sum across regions. SSIM~\cite{wang2004ssim} preserves structure and the VGG feature distance follows perceptual objectives~\cite{johnson2016perceptual,zhang2018lpips,simonyan2015vgg}. $\mathcal{L}_G$ is the angular difference between the requested direction and the output of a fixed gaze estimator, following prior redirection protocols~\cite{Ruzzi2022GazeNeRF,Wei2025GazeGaussian}. Observed pairs use all terms; intermediate samples provide gaze supervision without pretending that an unobserved pixel target is available.

\section{Experiment 1: Gaze-redirection evaluation}
\label{sec:experiments}

\begin{table*}[t]
\centering
\caption{Within-dataset redirection on the ETH-XGaze person-specific test set. All methods use the same training split and source--target pairs. Best results are bold.}
\label{tab:within}
\small
\setlength{\tabcolsep}{5.2pt}
\begin{tabular}{lccccccc}
\toprule
Method & Gaze $\downarrow$ & Pose $\downarrow$ & SSIM $\uparrow$ & PSNR $\uparrow$ & LPIPS $\downarrow$ & FID $\downarrow$ & ID $\uparrow$ \\
\midrule
STED & 16.217 & 13.153 & 0.726 & 17.530 & 0.300 & 115.020 & 24.347 \\
GazeNeRF & 6.944 & 3.470 & 0.733 & 15.453 & 0.291 & 81.816 & 45.207 \\
GazeGaussian & 6.622 & \textbf{2.128} & 0.823 & 18.734 & 0.216 & 41.972 & 67.749 \\
\textbf{DiTGaze} & \textbf{6.353} & 2.349 & \textbf{0.850} & \textbf{20.512} & \textbf{0.187} & \textbf{38.319} & \textbf{71.724} \\
\bottomrule
\end{tabular}
\end{table*}

\paragraph{Data.}
We train on ETH-XGaze~\cite{xu2020ethxgaze} and first evaluate on its person-specific test set. We follow the standardized GazeNeRF protocol~\cite{Ruzzi2022GazeNeRF}, also used by GazeGaussian~\cite{Wei2025GazeGaussian}: $512\!\times\!512$ resolution, 3D camera normalization, head-relative yaw--pitch labels, and the ETH-XGaze split. The common training subset contains 80 identities, 10 frames per view, and 18 views (14,400 images). Testing uses 15 unseen identities with 200 labeled images each and the same source--target pairing for all methods. Identities are disjoint across training and evaluation; per-person calibration frames are disjoint from evaluated pairs. Cross-dataset tests train only on ETH-XGaze and evaluate ColumbiaGaze~\cite{smith2013gazelocking}, MPIIFaceGaze~\cite{zhang2015mpiifacegaze,zhang2016fullfacegaze}, and GazeCapture~\cite{krafka2016eyetracking}. Dataset-specific camera normalization follows the published baseline preprocessing, and the same face parser~\cite{zllfaceparsing} separates eye and remaining-face supervision.

\paragraph{Baselines and metrics.}
We compare with the 2D latent redirector STED~\cite{Zheng2020STED}, the two-stream 3D neural renderer GazeNeRF~\cite{Ruzzi2022GazeNeRF}, and the 3DGS system GazeGaussian~\cite{Wei2025GazeGaussian}. Redirection is measured by gaze and head-pose angular error. Image assessment includes SSIM~\cite{wang2004ssim}, PSNR, LPIPS~\cite{zhang2018lpips}, FID~\cite{heusel2017fid}, and face-embedding identity similarity (ID), matching the strongest baseline's protocol.

\paragraph{Optimization and compute.}
We use Adam~\cite{kingma2015adam} with initial learning rate $10^{-4}$ and step decay. Geometry is initialized first; the complete renderer, Gaussians, and deformation fields are then optimized end-to-end while the latent VAE remains frozen. All experiments run in PyTorch on one NVIDIA A100-SXM4. Two-stage training takes approximately 110 GPU-hours and peaks at 14.2 GB; inference uses 4.8 GB. A one-time generator identity adaptation for an unseen subject takes approximately 30 s, after which the complete generation pipeline takes 165--190 ms per image (about 40 GFLOPs). For context, GazeGaussian reports 74 FPS for rendering in its main paper but approximately 0.2 s per image for end-to-end data generation in its supplement, plus a comparable 30 s generator identity adaptation, on an RTX 4090~\cite{Wei2025GazeGaussian}. The hardware and timing boundaries differ, so this is an order-of-magnitude comparison rather than a speedup claim: both systems occupy a similar practical offline-generation regime.

\subsection{Within-dataset redirection quality}

\begin{figure}[tb]
\centering
\includegraphics[width=0.98\linewidth]{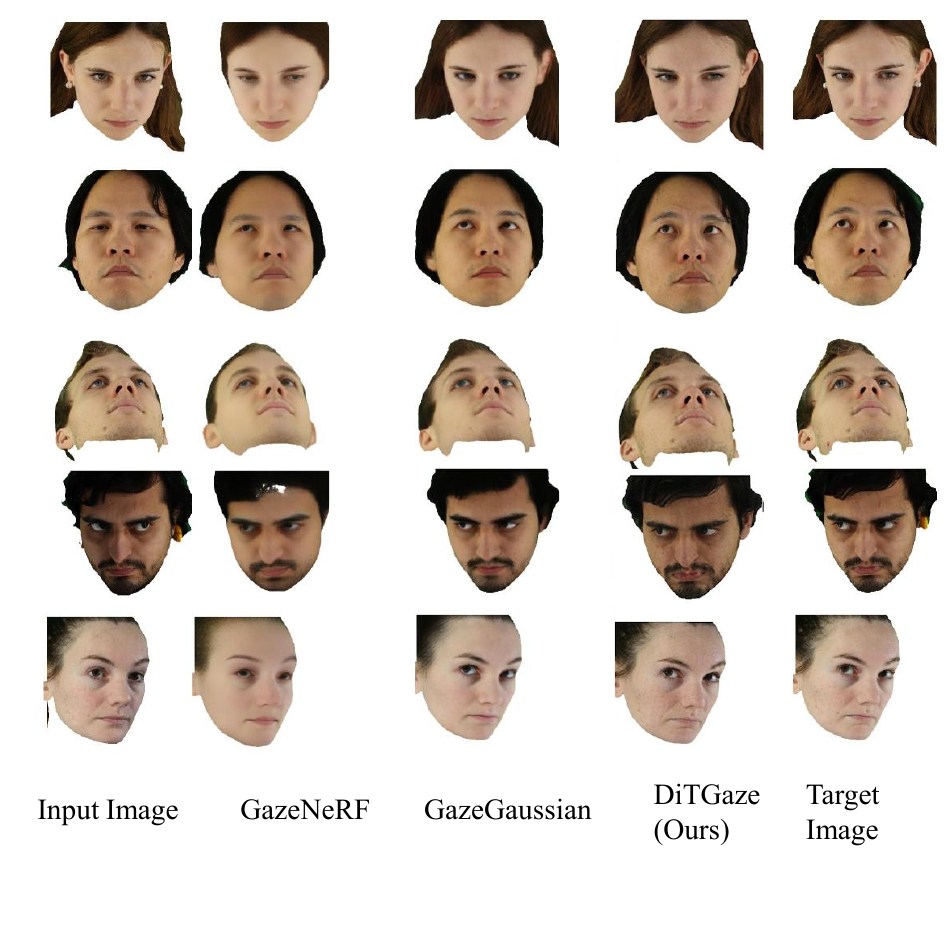}
\caption{Within-dataset qualitative comparison on ETH-XGaze. Columns show the input, prior 3D redirectors, \method, and target.}
\label{fig:qualitative}
\end{figure}

\Cref{tab:within} evaluates the within-dataset generator-level task. Relative to GazeGaussian, \method reduces gaze error from $6.622\degrees$ to $6.353\degrees$ (4.1\%), increases PSNR by 1.778 dB, reduces LPIPS by 13.4\%, and raises identity similarity by 3.975 points. These complementary metrics diagnose the label and appearance quality of generated supervision; \Cref{sec:downstream} tests its task-level utility. \Cref{fig:qualitative} shows sharper eyelids, irises, skin texture, and face boundaries than the deterministic baselines.

Head-pose error increases by $0.221\degrees$ relative to GazeGaussian, but remains $1.121\degrees$ lower than GazeNeRF. This is a measurable limitation rather than a uniformly dominant result: diffusion refinement improves perceptual quality and gaze accuracy while slightly relaxing pose rigidity. Removing orthogonality raises pose error further, from $2.349\degrees$ to $2.572\degrees$, indicating that the constraint mitigates rather than eliminates this trade-off.

\subsection{Cross-dataset generalization}

\begin{table*}[t]
\centering
\caption{Cross-dataset redirection after training on ETH-XGaze. We report angular gaze/pose error ($\degrees$), LPIPS, and identity similarity. Best results are bold.}
\label{tab:cross}
\small
\setlength{\tabcolsep}{8pt}
\begin{tabular}{llcccc}
\toprule
Dataset & Method & Gaze $\downarrow$ & Pose $\downarrow$ & LPIPS $\downarrow$ & ID $\uparrow$ \\
\midrule
\multirow{4}{*}{ColumbiaGaze}
& STED & 17.887 & 14.693 & 0.413 & 6.384 \\
& GazeNeRF & 9.464 & 3.811 & 0.352 & 23.157 \\
& GazeGaussian & 7.415 & \textbf{3.332} & 0.273 & 59.788 \\
& \textbf{DiTGaze} & \textbf{7.265} & 3.611 & \textbf{0.231} & \textbf{64.963} \\
\midrule
\multirow{4}{*}{MPIIFaceGaze}
& STED & 14.796 & 11.893 & 0.288 & 10.677 \\
& GazeNeRF & 14.933 & 7.118 & 0.272 & 30.981 \\
& GazeGaussian & 10.943 & \textbf{5.685} & 0.224 & 41.505 \\
& \textbf{DiTGaze} & \textbf{10.512} & 5.905 & \textbf{0.196} & \textbf{44.478} \\
\midrule
\multirow{4}{*}{GazeCapture}
& STED & 15.478 & 16.533 & 0.271 & 6.808 \\
& GazeNeRF & 10.463 & 9.064 & 0.232 & 20.981 \\
& GazeGaussian & 9.752 & \textbf{7.061} & 0.209 & 19.025 \\
& \textbf{DiTGaze} & \textbf{9.676} & 7.217 & \textbf{0.181} & \textbf{22.236} \\
\bottomrule
\end{tabular}
\end{table*}

The cross-dataset results in \Cref{tab:cross} test the redirector on unseen identities, cameras, and illumination. \method obtains the lowest gaze and LPIPS errors and highest identity similarity on all three datasets. The largest identity gain over GazeGaussian is 5.175 points on ColumbiaGaze; LPIPS falls from 0.273 to 0.231. Pose retains the same small trade-off observed within ETH-XGaze. The consistency indicates that the generator gains are not confined to ETH-XGaze appearance statistics; downstream transfer is evaluated separately in \Cref{sec:downstream}.

\subsection{Component ablation}
\label{sec:ablation}

\begin{table}[t]
\centering
\caption{Component ablation on ETH-XGaze. Each row removes one component from the full system.}
\label{tab:ablation}
\footnotesize
\setlength{\tabcolsep}{3.5pt}
\begin{tabular}{lcccc}
\toprule
Variant & Gaze $\downarrow$ & Pose $\downarrow$ & LPIPS $\downarrow$ & FID $\downarrow$ \\
\midrule
w/o DiT renderer & 6.485 & \textbf{2.328} & 0.214 & 41.878 \\
w/o orthogonality & 6.396 & 2.572 & 0.191 & 39.337 \\
w/o gaze sampling & 6.714 & 2.344 & 0.190 & 38.582 \\
\textbf{Full DiTGaze} & \textbf{6.353} & 2.349 & \textbf{0.187} & \textbf{38.319} \\
\bottomrule
\end{tabular}
\end{table}

\Cref{tab:ablation} reports the generator-level ablation; \Cref{tab:downstream_ablation} provides its end-to-end counterpart. Replacing DiT with the convolutional renderer raises LPIPS by 14.4\% and FID by 9.3\%, confirming that the renderer drives fidelity rather than eye geometry. Removing gaze sampling raises gaze error by 5.7\% to $6.714\degrees$, while removing orthogonality raises pose error by 9.5\% ($2.349\degrees\!\rightarrow\!2.572\degrees$). Qualitative ablations are provided in the supplement.

\section{Experiment 2: Downstream few-shot gaze estimation}
\label{sec:downstream}

Following GazeNeRF and GazeGaussian~\cite{Ruzzi2022GazeNeRF,Wei2025GazeGaussian}, we rerun the full protocol by fine-tuning an ETH-XGaze-trained ResNet-50~\cite{He2016ResNet} for 15 unseen subjects using $N\in\{1,\ldots,9\}$ real calibration images. Augmented sets contain $B=200$ samples; real-only uses just the $N$ user-provided observations and is therefore a practical calibration-budget reference, not a sample-count-matched control. \Cref{tab:downstream_curve} covers all nine budgets, while \Cref{tab:downstream_stats} repeats the primary inferential comparison at $N=9$ over five calibration draws and three estimator seeds. Complete held-out evaluation and aggregation details are provided in the supplement.

\begin{table}[t]
\centering
\caption{\textbf{Rerun of the ETH-XGaze few-shot curve.} Mean gaze error ($\degrees$) on held-out real images; augmented sets contain $B=200$ samples. Lower is better, and the best result at each calibration budget is bold.}
\label{tab:downstream_curve}
\footnotesize
\setlength{\tabcolsep}{3.2pt}
\begin{tabular}{c@{\hspace{4pt}}ccccc}
\toprule
$N$ real & Real & STED & GazeNeRF & GazeGaussian & \method \\
\midrule
1 & 9.1 & 8.8 & 6.9 & 6.1 & \textbf{5.7} \\
2 & 9.0 & 8.5 & 6.5 & 6.0 & \textbf{5.3} \\
3 & 8.8 & 9.0 & 6.4 & 6.0 & \textbf{5.4} \\
4 & 8.0 & 8.9 & 6.0 & 5.9 & \textbf{5.1} \\
5 & 6.8 & 8.0 & 6.4 & 5.8 & \textbf{5.2} \\
6 & 6.9 & 8.2 & 6.3 & 5.7 & \textbf{5.1} \\
7 & 6.7 & 8.1 & 6.1 & 5.5 & \textbf{4.9} \\
8 & 6.8 & 8.1 & 6.0 & 5.4 & \textbf{4.9} \\
9 & 6.7 & 7.9 & 5.9 & 5.3 & \textbf{4.8} \\
\bottomrule
\end{tabular}
\end{table}

\begin{table}[t]
\centering
\caption{\textbf{Primary paired comparison at $N=9$.} Results aggregate five draws and three seeds per subject; $\Delta$ is GazeGaussian minus \method, so positive values favor ours.}
\label{tab:downstream_stats}
\footnotesize
\setlength{\tabcolsep}{4pt}
\begin{tabular}{lcc}
\toprule
Method/statistic & Mean $\pm$ SD ($\degrees$) & 95\% CI ($\degrees$) \\
\midrule
GazeGaussian & $5.63 \pm 1.39$ & $[4.96,\ 6.33]$ \\
\textbf{DiTGaze} & $\mathbf{4.90 \pm 1.29}$ & $\mathbf{[4.29,\ 5.57]}$ \\
Paired $\Delta$ & $\mathbf{0.73 \pm 0.49}$ & $\mathbf{[0.49,\ 0.97]}$ \\
\midrule
Wilcoxon $p$ / effect size & \multicolumn{2}{c}{$p<0.001$ / $r_{\mathrm{rb}}=0.967$} \\
\bottomrule
\end{tabular}
\end{table}

\begin{table}[!b]
\centering
\caption{\textbf{Downstream component ablation} at ETH-XGaze $N=9$, $S=191$, with fixed real samples, targets, estimator, and fine-tuning schedule.}
\label{tab:downstream_ablation}
\footnotesize
\setlength{\tabcolsep}{4pt}
\begin{tabular}{lcc}
\toprule
Generator variant & Error ($\degrees$) $\downarrow$ & $\Delta$ vs. full \\
\midrule
w/o DiT renderer & 5.220 & +0.346 \\
w/o orthogonality & 4.995 & +0.120 \\
w/o gaze sampling & 5.555 & +0.680 \\
\textbf{Full DiTGaze} & \textbf{4.875} & 0.000 \\
\bottomrule
\end{tabular}
\end{table}

\begin{table*}[t]
\centering
\caption{\textbf{Calibration and synthetic-data sensitivity on ETH-XGaze.} Panel (a) varies the real calibration budget with $B=200$ augmented samples. Panel (b) independently fixes $N=9$ while varying synthetic count, so total size changes. Each panel is an internally paired controlled sweep.}
\label{tab:sensitivity}
\footnotesize
\begin{minipage}[t]{0.45\textwidth}
\centering
\textbf{(a) Real-shot curve}\par\smallskip
\begin{tabular}{cccc}
\toprule
$N$ & Real only & GazeGaussian & \method \\
\midrule
5  & 6.883 & 5.855 & \textbf{5.253} \\
9  & 6.704 & 5.439 & \textbf{4.906} \\
15 & 6.107 & 5.273 & \textbf{4.701} \\
\bottomrule
\end{tabular}
\end{minipage}
\hfill
\begin{minipage}[t]{0.50\textwidth}
\centering
\textbf{(b) Synthetic-count curve}\par\smallskip
\begin{tabular}{ccccc}
\toprule
$S$ & Total & Synth. \% & GazeGaussian & \method \\
\midrule
50  & 59  & 84.7 & 5.777 & \textbf{5.373} \\
100 & 109 & 91.7 & 5.439 & \textbf{4.906} \\
191 & 200 & 95.5 & 5.309 & \textbf{4.576} \\
300 & 309 & 97.1 & 5.282 & \textbf{4.556} \\
\bottomrule
\end{tabular}
\end{minipage}
\end{table*}

\begin{table*}[t]
\centering
\caption{\textbf{Downstream domain evaluation at $N=9$.} Mean angular error on held-out real images. ETH$\rightarrow$ETH is the in-domain reference; ETH$\rightarrow$MPII transfers the ETH-XGaze-pretrained estimator and generator before target-subject generator adaptation and estimator fine-tuning. MPII denotes MPIIFaceGaze.}
\label{tab:downstream_domain}
\footnotesize
\setlength{\tabcolsep}{6pt}
\begin{tabular}{lcc}
\toprule
Fine-tuning data & ETH$\rightarrow$ETH & ETH$\rightarrow$MPII \\
\midrule
Real only & 6.7 & 11.8 \\
Real + STED & 7.9 & 12.1 \\
Real + GazeNeRF & 5.9 & 10.4 \\
Real + GazeGaussian & 5.3 & 9.9 \\
Real + \textbf{DiTGaze} & \textbf{4.8} & \textbf{9.5} \\
\bottomrule
\end{tabular}
\end{table*}

\Cref{tab:downstream_curve} shows that \method ranks first from one to nine shots, reducing error by 10.3\% on average relative to GazeGaussian and by 31.8\% relative to real-only fine-tuning. STED is worse than real only from $N=3$, showing that synthetic volume alone is insufficient. At $N=9$, \method reaches $4.8\degrees$, compared with $5.3\degrees$ for GazeGaussian, $5.9\degrees$ for GazeNeRF, $7.9\degrees$ for STED, and $6.7\degrees$ for real only.

\paragraph{Repeated paired comparison.}
Matched $N=9$ repetitions lower subject-level error from $5.63\degrees$ to $4.90\degrees$, a paired reduction of $0.73\degrees$ (13.0\%; 95\% BCa CI $[0.49\degrees,0.97\degrees]$). \method improves 14 of 15 identities; the exact two-sided Wilcoxon test gives $W=2.0$, $p=1.83\!\times\!10^{-4}$, and $r_{\mathrm{rb}}=0.967$. These means differ from the full-curve $N=9$ estimate because \Cref{tab:downstream_stats} aggregates five draws and three seeds.

\subsection{Downstream mechanisms and budgets}

\Cref{tab:downstream_ablation} holds the real samples, target directions, estimator, and fine-tuning schedule fixed while changing only the generator variant. Every removal increases downstream error: omitting orthogonality, the DiT renderer, and gaze sampling costs $0.120\degrees$, $0.346\degrees$, and $0.680\degrees$, respectively. The largest penalty from removing gaze sampling agrees with its generator-level gaze-error effect in \Cref{tab:ablation}; the DiT penalty shows that its fidelity gains also survive estimator fine-tuning.

\Cref{tab:sensitivity} separates two deployment questions: varying $N$ measures required user effort, whereas varying $S$ measures the effect of synthetic-data volume at fixed real input. In the real-shot sweep, \method remains best at $N=5,9,15$ and improves from $5.253\degrees$ to $4.701\degrees$ as more real observations are supplied; its advantage over GazeGaussian remains $0.53$--$0.60\degrees$. In the independent synthetic-count sweep, increasing $S$ from 50 to 191 lowers \method's error by $0.797\degrees$, but increasing it further to 300 yields only $0.020\degrees$. This plateau suggests that extra samples increasingly revisit already covered yaw--pitch regions, so target diversity rather than raw count becomes limiting. It does not establish that every possible gaze label is exhausted; an independently labeled angular-coverage audit would be needed for that claim. Full nested-sampling and optimization controls are given in the supplement.

\subsection{In- and cross-domain downstream evaluation}

\Cref{tab:downstream_domain} separates in-domain personalization from cross-domain use. ETH$\rightarrow$MPII is cross-domain few-shot personalization rather than zero-shot evaluation because nine labeled target-person images are used. Domain shift raises every error, yet \method remains best at $9.5\degrees$, compared with $9.9\degrees$ for GazeGaussian, $10.4\degrees$ for GazeNeRF, $11.8\degrees$ for real-only fine-tuning, and $12.1\degrees$ for STED. The $2.3\degrees$ reduction relative to real only shows that redirected supervision remains useful after transfer, while the gap from $4.8\degrees$ in-domain to $9.5\degrees$ cross-domain leaves substantial room for domain-robust personalization. This downstream test is distinct from the generator-level cross-dataset evaluation in \Cref{tab:cross}.

\subsection{Cost, scope, and limitations}

The controlled ablations show that the gains do not arise from diffusion alone: target coverage and deformation regularization make separate contributions. The experiments do not directly establish monotonic angular response at withheld directions; an angular holdout audit with an independent gaze labeler would address that question.

Generating 191 samples takes 31.5--36.3 s after adaptation; hardware differences preclude speed ranking, and both systems target offline personalization. Thin accessories, extreme gaze, unusual lighting, the $0.221\degrees$ pose trade-off, and possible shared gaze-labeler bias remain limitations.

All experiments use public datasets and collect no new personal data. Deployments should minimize identity storage, record synthetic provenance, and audit demographic and device performance. Additional failure cases and evaluation-scope limitations are discussed in the supplement.

\section{Conclusion}

Accurate, low-burden personalization matters for engagement-aware e-learning, assistive systems, and interactive interfaces. \method combines explicit 3D Gaussian gaze control with a latent DiT renderer, yaw--pitch supervision, and deformation regularization. It reduces generator-level gaze error by 4.1\% relative to GazeGaussian and ranks first throughout the rerun one-to-nine-shot curve. At $N=9$, the repeated comparison reduces subject-level error by $0.73\degrees$ and improves 14 of 15 identities; \method also remains best under ETH$\rightarrow$MPII transfer. Gains saturate beyond 191 synthetic samples, indicating that angular coverage matters more than unbounded volume. Limitations include offline generation, a small pose trade-off, failures under thin accessories and extreme conditions, and substantial residual cross-domain error. Future work will explore faster personalization and stronger cross-domain robustness.

\clearpage
{\small
\bibliographystyle{ieeenat_fullname}
\bibliography{main}
}

\end{document}


\maketitle

\section{Scope}

This supplement provides implementation and diagnostic evidence for the main paper. \Cref{sec:supp_impl} expands the preprocessing, architecture, optimization, and inference protocol. \Cref{sec:supp_metrics} specifies the evaluation measures. \Cref{sec:supp_qual} provides additional qualitative and ablation examples. \Cref{sec:supp_failure,sec:supp_calib,sec:supp_limits} document failure cases, generator identity-adaptation behavior, and evaluation scope. No additional dataset result or improved version of the method is introduced here.

\section{Implementation Details}
\label{sec:supp_impl}

\subsection{Preprocessing}

Inputs are resized to $512\!\times\!512$. We obtain face and eye masks using the face-parsing model of~\cite{zllfaceparsing}; the mask separates eye-region and remaining-face Gaussians. Identity, expression, camera, and initial geometry follow the tracking pipeline used by Gaussian Head Avatar~\cite{xu2024gaussian}. Gaze labels are converted to head-relative yaw and pitch under the normalized camera convention used by GazeNeRF~\cite{Ruzzi2022GazeNeRF} and GazeGaussian~\cite{Wei2025GazeGaussian}.

Canonical geometry begins from a neutral signed-distance representation. DMTet~\cite{Shen2021DMTet} extracts the neutral mesh, and 3D landmarks partition it into face and eye regions. The resulting vertices initialize separate Gaussian sets. Each Gaussian contains position, feature, opacity, rotation, and scale parameters. Face Gaussians receive expression and pose deformation; eye Gaussians receive expression deformation and the explicit gaze-conditioned rotation.

\subsection{DiT Architecture}

The renderer uses six Diffusion Transformer blocks with eight attention heads per block. A learnable $1\!\times\!1$ projection, optimized jointly with the renderer, maps the 32-channel rasterized Gaussian feature representation to a three-channel VAE-compatible representation. The frozen VAE from latent diffusion~\cite{Rombach2021LDM} therefore never receives the raw 32-channel feature map; it compresses the learned projection to a four-channel latent map. A convolutional PatchEmbed layer creates the token grid, and two-dimensional sinusoidal positional embeddings preserve spatial layout.

\begin{table*}[t]
\centering
\caption{Latent DiT block used in \method. The six-block stack operates on patch tokens; timestep and identity/expression conditioning enter through AdaLN.}
\label{tab:supp_arch}
\small
\begin{tabular}{p{0.20\textwidth}p{0.23\textwidth}p{0.47\textwidth}}
\toprule
Stage & Operation & Function \\
\midrule
Input normalization & AdaLN & Layer normalization with conditioning-dependent channel scale and shift \\
Global mixing & 8-head self-attention & Relates local eye changes to full-face geometry and appearance \\
Residual update & Gated residual connection & Adds the conditioned attention response to the token stream \\
Feature transformation & AdaLN + MLP & Applies conditioned nonlinear processing to every token \\
Output & Gated residual connection & Produces the next denoising block state \\
\bottomrule
\end{tabular}
\end{table*}

The diffusion timestep is embedded and combined with identity and expression codes. This conditioning predicts the scale, shift, and residual gates used by each block. A final projection maps tokens back to patch values, unpatchification reconstructs the latent map, and the frozen VAE decoder produces the RGB image. The projection, DiT, Gaussian parameters, and deformation MLPs are trainable; VAE weights remain fixed.

\subsection{Training}

Population-level training uses the same ETH-XGaze subset as the main experiments: 80 training identities, 10 frames from each of 18 views, for 14,400 images. We train for 30 epochs with Adam~\cite{kingma2015adam}, initial learning rate $10^{-4}$, and step decay every 1,000 iterations. Geometry is initialized in the first stage; the projection, DiT renderer, Gaussian attributes, and deformation fields are then optimized jointly while the VAE remains frozen. Reconstruction and orthogonality terms are active from the start; angular-gaze and VGG perceptual terms are introduced after initial stabilization. The image objective combines $\ell_1$, SSIM~\cite{wang2004ssim}, and perceptual feature distance~\cite{zhang2018lpips,simonyan2015vgg} for the eye, remaining-face, and full-head outputs. The functional gaze loss uses a fixed VGG-based estimator to measure angular error between the rendered gaze and the requested direction. For observed pairs, the target comes from the paired image; for intermediate samples, it is the sampled yaw--pitch vector and no unobserved pixel target is imposed. The deformation-space orthogonality loss is added using the weighting terms defined in the main paper.

The Intermediate Gaze Sampler supports uniform, center-biased, and mixed distributions. The reported model uses progressive mixed sampling: early epochs emphasize frontal angles, followed by wider grid and random coverage. Grid points ensure deterministic coverage of the configured yaw--pitch range; random points expose between-cell directions and reduce lattice-specific fitting. The schedule avoids destabilizing geometry before the renderer has learned the person's coarse appearance while still exposing the final model to the full configured range.

Training uses one NVIDIA A100-SXM4 and takes about 110 GPU-hours over two stages. Peak memory is 14.2 GB during training and 4.8 GB during inference. GazeGaussian reports an RTX 4090, so the figures do not support an equal-hardware speed ranking.

\subsection{Unseen-Identity Generator Adaptation and Generation}

For a previously unseen subject, a short one-time generator identity-adaptation step updates only identity-specific Gaussian attributes, deformation parameters, and the identity latent code. The global DiT renderer and diffusion weights remain fixed. All reported experiments use approximately 30 s of optimization. The calibration frames are disjoint from the source--target evaluation pairs.

After generator adaptation, each requested gaze direction is generated through the complete rasterization and denoising pipeline in 165--190 ms per image on the A100-SXM4. GazeGaussian reports 74 FPS for rendering, but its supplementary end-to-end data-generation estimate is approximately 0.2 s per image and it uses a similarly brief, approximately 30 s generator identity adaptation on an RTX 4090~\cite{Wei2025GazeGaussian}. Thus, DiTGaze is not separated from the baseline by an order-of-magnitude practical generation penalty; the hardware difference precludes a direct winner claim. The resulting data are intended for offline augmentation.

\subsection{Downstream Evaluation Protocol}

We use the ResNet-50 gaze/head-pose estimator released with GazeNeRF~\cite{Ruzzi2022GazeNeRF}, trained on the ETH-XGaze training identities, as the common initialization for every downstream condition. Real and generated face patches receive the same $224\!\times\!224$ estimator preprocessing. The person-specific set contains 15 identities excluded from base-estimator training: 0012, 0022, 0025, 0047, 0054, 0074, 0086, 0094, 0096, 0097, 0110, 0113, 0115, 0116, and 0119. Each identity has 200 labeled real images. Fifteen images form a calibration-candidate pool and the remaining 185 form the fixed held-out evaluation set. Five random orderings of the calibration pool define the calibration draws; an $N$-shot condition uses the first $N$ images in that ordering, while unused calibration candidates are excluded from both fine-tuning and evaluation.

For each augmented condition, the $N$ real images seed the redirector and $S=200-N$ images are generated, giving exactly 200 fine-tuning samples. The 191 synthetic images at $N=9$ are generated in total across the nine real sources, not per source. Real-only models use only the $N$ unique calibration images; the loader cycles through them for the fixed optimization budget without adding further unique observations. All methods share calibration indices, requested yaw--pitch targets, source assignments, held-out images, the initial estimator checkpoint, and the number of estimator updates.

For every held-out image, predicted and reference yaw--pitch values are converted to unit 3D gaze vectors. Angular error is $180/\pi$ times the arccosine of their clipped dot product; per-image angular errors are then averaged, rather than averaging gaze vectors first. The same coordinate convention is used for pre-training, fine-tuning, and evaluation, and held-out images are never used for checkpoint selection.

For the primary $N=9$ comparison, each generated pool is held fixed while the estimator is fine-tuned from the common checkpoint with three optimization seeds. We first average image-level angular error over the 185 held-out images, then average the three estimator seeds within each subject and draw, the five draws within each subject, and finally the 15 subject means. Standard deviation is reported across subject means. The 95\% intervals use 10,000 subject-level BCa bootstrap resamples, with paired resampling for the difference. The prespecified inferential comparison is an exact two-sided Wilcoxon signed-rank test at $\alpha=0.05$ on the 15 full-precision subject differences; matched-pairs rank-biserial correlation is the effect size. Images, draws, and optimization seeds are not treated as independent observations. The other budget and ablation sweeps are controlled descriptive analyses rather than additional confirmatory tests.

The remaining downstream tests preserve the same controls. Component ablations use $N=9$ and $S=191$, changing only the generator variant. The real-shot study uses nested $N\in\{5,9,15\}$ prefixes, while the independently executed synthetic-count study fixes $N=9$, generates one 300-image target pool per subject, and uses nested prefixes for $S\in\{50,100,191,300\}$; $S=0$ is real-only. Estimator update count is fixed across $S$. The retained domain experiment transfers the ETH-XGaze-pretrained estimator and generator to target-person adaptation and fine-tuning on MPIIFaceGaze; it remains few-shot rather than zero-shot because nine labeled target-person images are used. Immutable manifests record subject, calibration and held-out indices, target directions, source assignment, generator checkpoint, estimator seed, and nested-prefix membership for every condition.

\section{Metric Definitions}
\label{sec:supp_metrics}

\paragraph{Gaze and head-pose error.}
We compute the angular distance in degrees between the estimated 3D direction of a rendered image and its target. Gaze error measures label fidelity; head-pose error measures whether a gaze edit unintentionally changes pose.

\paragraph{Image fidelity.}
SSIM~\cite{wang2004ssim} and PSNR measure reconstruction similarity, LPIPS~\cite{zhang2018lpips} measures perceptual feature distance, and FID~\cite{heusel2017fid} compares generated and reference feature distributions. Identity similarity is computed from face-recognition embeddings under the same protocol as GazeGaussian~\cite{Wei2025GazeGaussian}. These metrics are complementary: none alone establishes useful synthetic supervision, which motivates the downstream gaze-estimation experiment in the main paper.

\section{Additional Qualitative Results}
\label{sec:supp_qual}

\begin{figure}[t]
\centering
\includegraphics[width=0.98\linewidth]{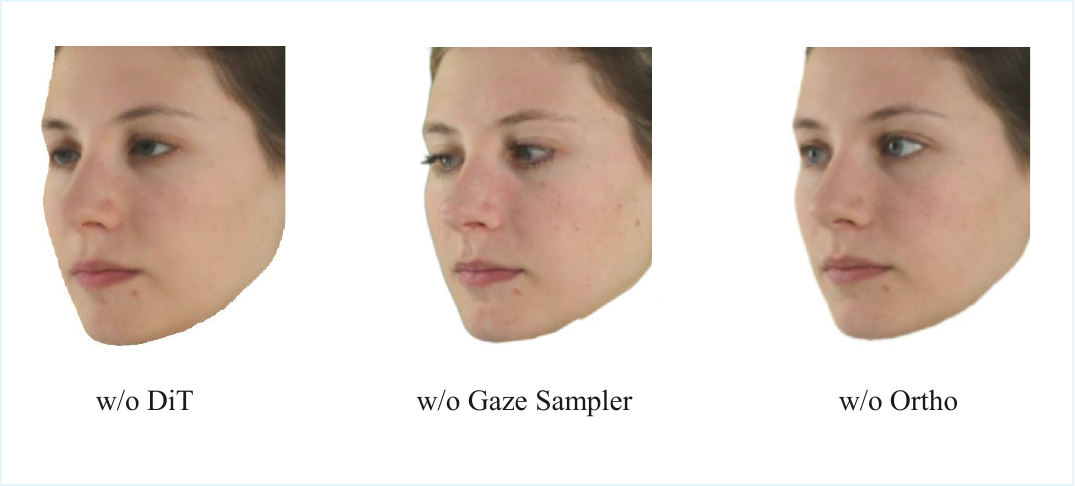}
\caption{Qualitative component ablations. From left to right: without DiT, without Intermediate Gaze Sampling, and without the orthogonality constraint.}
\label{fig:supp_ablation}
\end{figure}

\Cref{fig:supp_qual} provides the full controlled gaze-sweep visualization, placed in the supplement to preserve main-paper space for the architecture and application experiments. Each row holds subject identity and head configuration fixed while traversing requested gaze directions. The examples show that the renderer typically preserves face shape, hair boundaries, and local skin texture while moving the eyes toward the requested direction. \Cref{fig:supp_ablation} isolates characteristic component failures: removing DiT smooths high-frequency detail, removing the gaze sampler weakens subtle angular control, and removing orthogonality introduces cross-factor appearance inconsistency.

\begin{figure*}[t]
\centering
\includegraphics[width=0.96\textwidth]{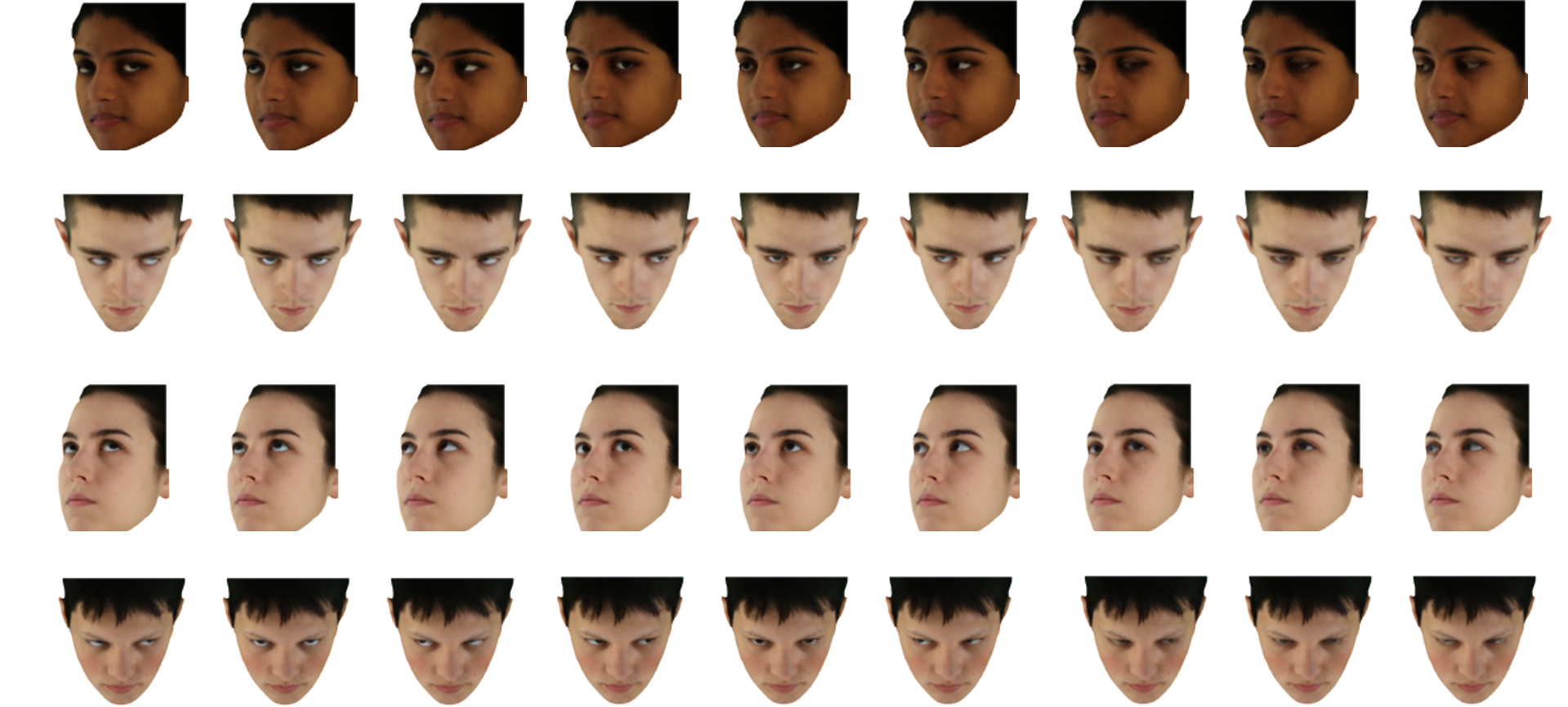}
\caption{Controlled multi-subject gaze sweeps on ETH-XGaze. Each row keeps the subject and head configuration fixed while traversing requested gaze directions. Smooth changes provide qualitative evidence complementary to angular-error evaluation; they are not used as a substitute for it.}
\label{fig:supp_qual}
\end{figure*}

\section{Failure Cases}
\label{sec:supp_failure}

\begin{figure}[t]
\centering
\includegraphics[width=0.23\linewidth]{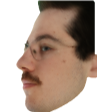}
\caption{Accessory failure: glasses are partly merged with facial appearance because the initial tracking and Gaussian geometry do not represent the thin structure reliably.}
\label{fig:supp_failure}
\end{figure}

\method inherits limitations from facial tracking, neutral-mesh initialization, and the finite Gaussian representation. Thin or partially occluding structures are difficult to reconstruct. In \Cref{fig:supp_failure}, glasses become blurred and partly merge with the skin after redirection. Similar artifacts can occur for earrings and fine hair. The diffusion renderer improves local texture in most cases but cannot restore geometry absent from the tracked representation.

Extreme gaze directions and lighting far outside the training distribution can also yield subtle stochastic inconsistencies. These failures are important for augmentation: visually plausible output is insufficient if the requested gaze is not retained. Training pipelines should filter generated samples with a separately trained gaze estimator not used by the generator loss, and retain the real calibration frames in every fine-tuning set.

\section{Generator Adaptation Behavior}
\label{sec:supp_calib}

Both GazeGaussian and \method use generator identity adaptation to update the Gaussian representation. Short optimization improves identity similarity and gaze accuracy by fitting subject-specific geometry and appearance. However, longer adaptation can over-specialize to the calibration views. Fine-grained skin texture, illumination, hair, and contour cues observed during adaptation may then persist in redirected outputs even when weakly supported by the current source view.

This effect can be more visible with global self-attention and repeated AdaLN conditioning because identity-specific information is distributed across tokens and layers. It is therefore a trade-off, not an unqualified benefit of longer optimization. We fixed generator identity adaptation at approximately 30 s for every reported experiment, kept evaluation pairs disjoint, and did not tune duration per test identity. Future work should report adaptation-duration curves and appearance-persistence measures alongside identity similarity.

\section{Evaluation Scope and Limitations}
\label{sec:supp_limits}

The synthetic-count plateau is consistent with the sampled yaw--pitch region approaching useful coverage, but it does not show that all possible gaze labels are exhausted or that response is monotonic between requested directions. A dedicated angular holdout with labels from an estimator independent of the generator loss would test both questions. The current gaze-fidelity metric may otherwise share some labeler bias with training, even though downstream evaluation uses held-out real images.

The ETH$\rightarrow$MPII result demonstrates cross-domain few-shot use, but its $9.5\degrees$ error remains substantially above the $4.8\degrees$ in-domain result. Broader camera, demographic, illumination, and device coverage is therefore needed before deployment claims. The system also performs generation offline, inherits tracking failures for thin accessories, and shows a $0.221\degrees$ head-pose trade-off relative to GazeGaussian. These limitations motivate domain-robust adaptation, independent angular validation, and more efficient generation rather than larger synthetic sets alone.

\section{Reproducibility Summary}

\begin{table}[h]
\centering
\caption{Key controls used across the reported comparisons.}
\footnotesize
\begin{tabular}{p{0.25\linewidth}p{0.60\linewidth}}
\toprule
Item & Reported control \\
\midrule
Training data & ETH-XGaze standard training split; identities disjoint from testing \\
Within-dataset evaluation & ETH-XGaze person-specific test set; common GazeNeRF source--target pairing \\
Cross-dataset evaluation & Train on ETH-XGaze only; test ColumbiaGaze, MPIIFaceGaze, and GazeCapture \\
Image resolution & $512\!\times\!512$ under the normalized 3D camera protocol \\
Downstream estimator & ResNet-50 pre-trained on ETH-XGaze training identities \\
Few-shot input & Same $N=9$ real calibration images for every augmentation method \\
Fine-tuning budget & Exactly 200 real-plus-synthetic samples per augmented condition \\
Repeated comparison & Five calibration draws $\times$ three estimator seeds; paired within subject \\
Estimator optimization & Identical optimizer, learning rate, batch size, data-order policy, and fixed update count across paired conditions; held-out images are not used for model selection \\
Statistical unit & Subject; seeds averaged within subject before paired inference \\
Generator identity adaptation & Approximately 30 s; evaluation pairs excluded; global renderer frozen \\
Hardware & One NVIDIA A100-SXM4 for \method training and evaluation \\
\bottomrule
\end{tabular}
\end{table}

\section{Per-Subject Statistical Results}

\Cref{tab:supp_subject_stats} decomposes the same $N=9$ runs summarized in the main paper's primary paired comparison; it does not introduce an additional experiment.

\Cref{tab:supp_subject_stats} makes the paired comparison auditable. Each entry is the mean over five calibration draws and three optimization seeds for that subject. The final row is computed over subjects, not over all images or runs pooled together.

\begin{table}[h]
\centering
\caption{Per-subject $N=9$ downstream error. Positive $\Delta$ favors \method.}
\label{tab:supp_subject_stats}
\footnotesize
\setlength{\tabcolsep}{4pt}
\begin{tabular}{cccc}
\toprule
Subject & GazeGaussian & \method & $\Delta$ \\
\midrule
0012 & 8.760 & 7.938 & 0.822 \\
0022 & 6.062 & 5.575 & 0.487 \\
0025 & 6.908 & 5.520 & 1.387 \\
0047 & 6.515 & 4.800 & 1.715 \\
0054 & 6.349 & 5.488 & 0.860 \\
0074 & 5.057 & 4.358 & 0.699 \\
0086 & 4.789 & 3.549 & 1.240 \\
0094 & 5.582 & 5.832 & -0.250 \\
0096 & 2.740 & 2.659 & 0.081 \\
0097 & 6.297 & 5.859 & 0.438 \\
0110 & 6.170 & 5.447 & 0.723 \\
0113 & 4.650 & 3.995 & 0.655 \\
0115 & 5.687 & 5.017 & 0.670 \\
0116 & 4.712 & 3.775 & 0.937 \\
0119 & 4.189 & 3.699 & 0.490 \\
\midrule
Mean $\pm$ SD & $5.631\pm1.387$ & $4.901\pm1.287$ & $0.730\pm0.487$ \\
\bottomrule
\end{tabular}
\end{table}

\FloatBarrier

{\small
\bibliographystyle{ieeenat_fullname}
\bibliography{main}
}